\documentclass[11pt]{article}

\usepackage[a4paper, margin=1in]{geometry}

\usepackage[T1]{fontenc}
\usepackage[utf8]{inputenc}
\usepackage{lmodern}
\usepackage{microtype}
\usepackage{textcomp}

\usepackage{amsmath, amssymb}

\usepackage{booktabs}
\usepackage{tabularx}
\usepackage{array}

\usepackage{enumitem}
\setlist{itemsep=2pt, topsep=4pt}

\usepackage[round]{natbib}
\usepackage{xcolor}
\usepackage[hidelinks]{hyperref}
\hypersetup{
  colorlinks=true,
  linkcolor=blue!50!black,
  citecolor=blue!50!black,
  urlcolor=blue!50!black
}

\usepackage{titlesec}
\titleformat{\section}{\large\bfseries}{\thesection.}{0.6em}{}
\titleformat{\subsection}{\normalsize\bfseries}{\thesubsection}{0.6em}{}

\title{\Large\bfseries Agent Manufacturing:\\
Foundation-Model Agents as First-Class Industrial Entities}

\author{
  Yilei Zhang \\[2pt]
  Department of Mechanical Engineering \\
  University of Canterbury \\
  Christchurch, New Zealand \\
  \texttt{yilei.zhang@canterbury.ac.nz}
}

\date{Preprint, May 2026}

\begin{document}

\maketitle

\begin{abstract}
\noindent
Manufacturing has passed through four widely recognized paradigms --- mechanization, electrification, programmable automation, and Smart Manufacturing --- each defined by the kind of work it shifted from humans to machines. In every case, one layer of industrial work remained fundamentally human: the \emph{coordinative cognition} of production, comprising the interpretive, allocative, diagnostic, negotiative, and governance work exercised by engineers, planners, and operational managers. We argue that a fifth transition is now underway in which this layer, rather than the physical or routine-cognitive layers below it, is what foundation-model-based autonomous agents primarily redistribute.

We name this paradigm \textbf{Agent Manufacturing} and define it operationally: a manufacturing system is an instance of Agent Manufacturing when its principal coordination mechanism is reasoning performed by foundation-model agents that can interpret open-ended goals, plan over long horizons, invoke tools and machines, and negotiate with other agents and humans. This is a narrower and more falsifiable definition than the existing literature on cognitive manufacturing or Industry 5.0 provides, and it distinguishes the paradigm sharply from classical multi-agent manufacturing systems, which were autonomous only within closed protocol spaces.

We make three further claims. First, the recent generation of vision-language-action models (RT-2, Open X-Embodiment, OpenVLA, $\pi_0$) and multi-agent orchestration frameworks (AutoGen, MetaGPT, industrial copilot deployments) constitutes the first technical substrate that could plausibly support Agent Manufacturing at scale, even though current evidence shows the technology is far from industrial reliability --- a 2025 systematic survey of foundation-model agents in industrial automation found 75\% of reported systems at technology readiness level 4--6 and only 9.1\% with deployment-oriented evidence. Second, the labor implications differ qualitatively from prior industrial revolutions: previous waves displaced muscle and routine cognition, whereas Agent Manufacturing displaces coordinative cognition, which has historically been the absorption layer for workers displaced by earlier waves and which Acemoglu and Restrepo's recent work identifies as a category of \emph{high-rent} labor whose automation amplifies wage losses. Third, the relevant unit of geopolitical competition is shifting from manufacturing capacity to \emph{cognitive manufacturing infrastructure} --- the foundation models, memory systems, and orchestration platforms on which industrial agents depend --- and this shift is now visible in concrete policy: the EU AI Act's high-risk classification regime, U.S. export controls on advanced chips and model weights, and China's ``AI+ Manufacturing'' action plan targeting 1{,}000 industrial intelligent agents by 2027.

We close with a manufacturing-specific research agenda, an explicit qualification that scaling current LLMs may not be the path to industrial-grade coordination, and a note that the resolution of these questions is not a technical matter alone.
\end{abstract}

\section{Introduction}
\label{sec:intro}

Each prior industrial revolution can be characterized by the kind of human work it displaced. Mechanization displaced muscle. Electrification reorganized the coordination of energy and motion. Programmable automation displaced routine motor and decision tasks on the shop floor. Smart Manufacturing --- under the labels Industrie 4.0 \citep{kagermann2013}, cyber-physical production systems \citep{monostori2014, lee2008cps}, and cloud manufacturing --- displaced certain forms of sensing, monitoring, and local optimization.

In every case, the work that remained centrally human was the coordinative cognition of production: interpreting what a customer or strategic goal actually requires, decomposing it into producible work, sequencing and rescheduling under disturbance, negotiating between conflicting objectives such as cost, quality, lead time, and sustainability, and exercising the judgment needed when written procedure runs out. Even highly automated factories today rely on this layer being supplied by people --- process engineers, planners, line leaders, quality engineers, supply-chain coordinators.

The persistence of this layer across four industrial revolutions is not accidental. Simon's classical analysis of organizations \citep{simon1947, simon1955} frames it directly: organizations exist as a structured response to bounded rationality, channeling decisions through hierarchies, rules, and communication channels precisely because individual cognition cannot encompass the complexity of production at scale. The coordinative layer in manufacturing is Simon's layer made concrete. It was protected from automation not because earlier technologies were too weak to perform optimization within it, but because they could not perform the open-ended interpretation, decomposition, and negotiation that the layer was built to manage.

The recent generation of foundation models and autonomous agents is the first technology that operates substantively above this boundary. Large language models can interpret ambiguous goals; agent frameworks can decompose them \citep{sumers2023}; vision-language-action models can ground plans in physical action \citep{brohan2023, kim2024openvla}; multi-agent orchestration frameworks can negotiate between sub-tasks \citep{wu2023autogen, hong2023metagpt}. None of these capabilities is yet reliable enough for unsupervised industrial deployment, and we will be explicit about this throughout the paper. We argue that they are the first capabilities of the right \emph{type} to displace coordinative cognition rather than merely physical or routine-cognitive work, and that this fact has consequences which the existing manufacturing literature has not adequately addressed.

This paper makes the case for treating this as a distinct paradigm --- Agent Manufacturing --- rather than as a continuation of Smart Manufacturing or a special case of multi-agent systems. We position the paradigm against three adjacent framings and against current industrial conditions (\S\ref{sec:adjacent}), give an operational definition and decompose the kinds of industrial cognition at stake (\S\ref{sec:definition}), distinguish it from classical industrial MAS (\S\ref{sec:mas}), describe the factory as a cognitive ecosystem with a worked example anchored in current deployments (\S\ref{sec:ecosystem}), examine the distinctive labor and political-economy consequences with reference to recent empirical work (\S\ref{sec:labor}, \S\ref{sec:politicaleconomy}), and outline a manufacturing-specific research agenda (\S\ref{sec:agenda}). Section~\ref{sec:conclusion} concludes with explicit attention to the epistemic limits of the argument.

\section{Why Agent Manufacturing Is Not Just Smart Manufacturing, Cognitive Manufacturing, or Industry 5.0}
\label{sec:adjacent}

Three adjacent framings already occupy nearby conceptual territory, and the claim that Agent Manufacturing is a distinct paradigm must be defended against each. We close this section by noting why current industrial conditions are making the workflow-centric architectures common to those framings increasingly costly to maintain, and provide a comparative table (Table~\ref{tab:paradigms}) summarizing the distinctions.

\textbf{Smart Manufacturing and Industrie 4.0} are built on cyber-physical integration, IIoT, digital twins, and data-driven optimization. They assume that the cognitive work of production --- what to make, how to schedule it, how to respond to a novel disturbance --- remains a human responsibility, supported by analytics. Their reference architectures (RAMI 4.0, IIRA) treat AI as a service invoked by human-designed workflows, not as an entity that designs workflows. Agent Manufacturing inverts this: workflows are increasingly generated, modified, and negotiated by agents under human governance, rather than the reverse.

\textbf{Cognitive Manufacturing}, as promoted in industry literature from the late 2010s onward (notably by IBM and consultancy reports), uses ``cognitive'' loosely --- typically to mean machine learning applied to quality, maintenance, and process optimization. The cognition referenced is statistical pattern recognition over industrial data. Agent Manufacturing makes a stronger claim: the relevant cognition is symbolic, goal-directed, and compositional, and it is performed by entities that maintain state, take action, and can be held accountable to objectives.

\textbf{Industry 5.0}, as articulated in EU policy documents and in academic syntheses \citep{xu2021industry5, leng2022industry5, xu2025embodied}, emphasizes human-centricity, sustainability, and resilience as values that should constrain industrial AI. Industry 5.0 is largely normative --- it specifies what industrial systems should be \emph{for}. Agent Manufacturing is descriptive --- it characterizes a structural change in \emph{how} such systems work. The two are compatible: one can pursue Industry 5.0's values within an Agent Manufacturing substrate, and indeed the human-governance questions raised in \S\ref{sec:labor} are clearer when posed in those terms.

The distinguishing feature of Agent Manufacturing is therefore neither cyber-physical integration (Industrie 4.0 already has it), nor the application of machine learning to industrial problems (Cognitive Manufacturing already does it), nor a values commitment to human centricity (Industry 5.0 already articulates it). The distinguishing feature is that the \emph{coordination mechanism itself} --- the process by which industrial work is interpreted, decomposed, allocated, and adapted --- is increasingly performed by foundation-model agents rather than by human planners or by fixed control logic.

\begin{table}[t]
\centering
\caption{Agent Manufacturing in relation to adjacent paradigms.}
\label{tab:paradigms}
\footnotesize
\renewcommand{\arraystretch}{1.25}
\begin{tabularx}{\textwidth}{@{}p{2.4cm}XXXXX@{}}
\toprule
\textbf{Dimension} & \textbf{Smart Mfg / Industrie 4.0} & \textbf{Classical Multi-Agent Mfg} & \textbf{Cognitive Mfg} & \textbf{Industry 5.0} & \textbf{Agent Manufacturing} \\
\midrule
Primary coordination locus & Human planning + predefined workflow & Predefined protocol (e.g., contract net) & Statistical analytics + human action & Human (by normative commitment) & Foundation-model-mediated coordination reasoning \\
Cognition type assumed & Optimization within fixed objectives & Symbolic, rule-based, bounded & Pattern recognition over industrial data & Augmented human judgment & Open-vocabulary, compositional, goal-directed \\
Autonomy boundary (\S\ref{sec:mas}) & Fixed workflow & Bounded protocol (thin autonomy) & Model inference scope & Human-constrained & Open-ended runtime reasoning (thick autonomy) \\
Locus of adaptation & Engineer redesign & Adaptation within bounded protocol space & Model retraining & Human--machine collaboration & Runtime agent reasoning \\
Reference architecture & RAMI 4.0, IIRA & PROSA, holonic mfg & Vendor-specific & EU policy frameworks & Emerging agent orchestration (e.g., CoALA-style, \citealp{sumers2023}) \\
Primary human role & Operator, supervisor & Protocol designer, exception handler & Analytics consumer & Central normative stakeholder & Governor of distributed cognition \\
Characteristic failure mode & Brittle workflow under disturbance & Protocol deadlock, narrow scope & Prediction drift, distributional shift & Governance ambiguity & Emergent coordination instability \\
Verification approach & Deterministic validation (IEC 61508, ISO 13849) & Protocol correctness proofs & Statistical evaluation, A/B testing & Process and ethics review & Open research problem (\S\ref{sec:agenda}) \\
Conceptual maturity & Operationally established (CPS + IIoT + DT) & Formally established (BDI, MAS) & Loosely defined, marketing-influenced & Primarily normative & Emerging operational definition (\S\ref{sec:definition}) \\
\bottomrule
\end{tabularx}
\end{table}

A clarification before continuing: the comparison in Table~\ref{tab:paradigms} should not be read as a claim that Agent Manufacturing displaces the prior paradigms. Industrial paradigms layer rather than replace --- a deployed Agent Manufacturing system runs on top of cyber-physical infrastructure inherited from Industrie 4.0, invokes analytics components characteristic of Cognitive Manufacturing, may coordinate with classical MAS schedulers, and is responsibly deployed only within governance constraints of the kind Industry 5.0 articulates. What changes is the dominant coordination layer, not the existence of the layers beneath it.

A reasonable response is that this distinction, even if real, matters only if existing paradigms are inadequate. We believe they increasingly are, for reasons grounded in current industrial conditions rather than in any inherent limitation of Smart Manufacturing as a framework. Supply-chain volatility since 2020 has repeatedly exceeded what static workflow optimization can absorb. The shift toward high-mix low-volume production --- driven by mass customization in consumer goods, personalization in medical devices, and the proliferation of regional regulatory variants --- multiplies the number of coordination decisions per unit of output. Energy price instability and decarbonization mandates require production schedules to respond to factors that traditional ERP systems were not designed to incorporate. Distributed and nearshored production multiplies the number of sites that must coordinate. In each case, the cost of having humans perform the coordinative work scales unfavorably, while the cost of having software perform it through fixed workflows scales with the brittleness of those workflows when conditions shift. Agent Manufacturing's appeal is not that workflow-centric automation is wrong; it is that the conditions under which workflow-centric automation was efficient are eroding.

\section{Definition and Decomposition of Industrial Cognition}
\label{sec:definition}

\subsection{Operational definition}

We define Agent Manufacturing operationally:

\begin{quote}
\itshape
A manufacturing system is an instance of Agent Manufacturing when its principal coordination mechanism is reasoning performed by foundation-model agents.
\end{quote}

We treat the following as necessary conditions for a system to qualify:

\begin{enumerate}
\item \textbf{Open-vocabulary goal interpretation.} The system can accept goals expressed in natural language or other open representations, without those goals having to be mapped in advance to a fixed schema.
\item \textbf{Long-horizon planning.} The system can decompose such goals into multi-step plans whose horizon exceeds the lookahead of any pre-specified control logic in the system.
\item \textbf{Tool and machine invocation.} Agents can invoke physical equipment, software services, or other agents as tools, including tools the agent was not specifically pre-trained on.
\item \textbf{Inter-agent negotiation.} Coordination among sub-tasks proceeds through agent-to-agent exchange rather than only through centralized scheduling.
\item \textbf{Memory and adaptation.} Agents maintain state across episodes such that behavior reflects accumulated experience rather than only current input.
\item \textbf{Human governance interface.} The system exposes mechanisms for human override, escalation, and audit at well-defined points.
\end{enumerate}

A system satisfying all six conditions is Agent Manufacturing in the strict sense. Systems satisfying a subset are partial or transitional instances; we expect most early deployments to fall in this category, and the systematic survey by \citet{henkel2026fmagents} confirms that current systems satisfy these conditions only partially, with most concentrated in user assistance, monitoring, and process optimization rather than in full coordinative roles.

We deliberately exclude conditions about \emph{what} the agents are coordinating: Agent Manufacturing can apply to discrete production, process industries \citep{ren2025industrialfm}, additive manufacturing, or supply chains, and to physical, informational, or hybrid tasks. The definition is intended to be falsifiable. A reader should be able to inspect a deployed system and decide whether it qualifies. A purely LLM-augmented MES that maps natural-language requests onto a fixed set of pre-coded actions, for example, fails condition~1 and probably condition~2; it is \emph{adjacent} to Agent Manufacturing without being an instance.

\subsection{What ``industrial cognition'' actually refers to}

The phrase ``industrial cognition'' recurs throughout this paper and risks becoming rhetorically powerful but analytically vague unless decomposed. We propose a functional decomposition that maps to recognizable industrial roles and that allows differentiated claims about which kinds of cognition current systems can plausibly perform.

\textbf{Interpretive cognition} translates external requirements --- customer needs, strategic objectives, regulatory constraints --- into producible work. It is exercised by product engineers, application engineers, and design-for-manufacturability specialists. Current LLM-based design copilots can perform parts of this work for well-precedented products \citep[as in Siemens Industrial Copilot deployments at Thyssenkrupp and similar,][]{siemens2024copilot}, but reliable interpretation across novel constraint combinations remains an open problem.

\textbf{Allocative cognition} assigns work to resources and sequences it over time. It is the cognition of production planning, scheduling, and dispatch. Classical multi-agent manufacturing systems and contract-net architectures already perform parts of this work within fixed protocol spaces; Agent Manufacturing extends it to cases where the protocol itself must be modified at runtime.

\textbf{Diagnostic cognition} identifies the causes of disturbances and failures. It is the abductive reasoning of quality engineers, maintenance specialists, and process troubleshooters. Foundation-model agents can perform parts of it in restricted laboratory settings \citep[in robotic contexts]{huang2022zeroshot, ahn2022saycan}, but industrial deployment requires reliability that current systems do not approach.

\textbf{Negotiative cognition} resolves conflicts between objectives, stakeholders, or sub-systems. It is exercised in cross-functional coordination, supplier management, and prioritization under contention. This is the kind of cognition most weakly supported by current agent frameworks; even where agents can negotiate within structured games, open-ended industrial negotiation remains research.

\textbf{Governance cognition} sets and revises objectives, calibrates trust in subordinate systems, and audits outcomes. It is the cognition of management. We will argue throughout this paper that this category should remain primarily human in any responsibly deployed Agent Manufacturing system, regardless of technical capability --- for reasons of accountability, not capability.

This decomposition does not exhaust industrial cognition (we have set aside, for example, creative design cognition and tacit craft knowledge), but it covers the categories that Agent Manufacturing most directly affects. Throughout the rest of the paper, when we say ``industrial cognition,'' we refer to this set.

\section{Distinction from Classical Multi-Agent Manufacturing Systems}
\label{sec:mas}

Multi-agent manufacturing has a substantial prior literature: holonic manufacturing systems, contract-net-based shop-floor control, and a body of work surveyed in \citet{shen2006mas} and grounded in \citet{wooldridge2009}. A reasonable objection to Agent Manufacturing is that it is simply this literature relabeled. It is not, and the distinction is worth making precisely because the architectures look superficially similar --- and indeed because recent work is actively combining the two, for example by inserting LLMs into the reasoning layers of holonic architectures \citep{ashfaq2025holonicllm}.

Classical industrial MAS agents are autonomous within a \emph{closed} coordination space. The set of message types, the ontology, the negotiation protocol (e.g., contract net), and the optimization objective are all specified in advance by the system designer. The agent's autonomy is the autonomy to make \emph{which choice} within that pre-specified space --- to bid or not to bid, to accept or reject. This is what we will call \emph{thin autonomy}: independence of decision within a fixed protocol.

Foundation-model agents in the sense relevant here exhibit \emph{thick autonomy}: the protocol itself is partially generated, modified, or extended at runtime by the agent's reasoning. An agent can express a coordination need that was not in its designer's ontology, propose a workaround for a tool that has failed in a novel way, or recognize that a sub-goal has become inappropriate given new information and renegotiate it. This capability is not new in kind to AI research (it has antecedents in planning, BDI architectures, and more recent cognitive architectures for language agents \citep{sumers2023}), but it is new in \emph{practical scope}: foundation models make open-vocabulary, open-domain reasoning cheap enough and broadly enough applicable to begin deploying in industrial settings.

This distinction matters because the safety, governance, and certification problems posed by thin-autonomy and thick-autonomy systems are fundamentally different. The classical industrial MAS literature largely solved its safety problems by bounding the protocol space. That solution does not transfer. Hybrid architectures that combine an LLM-driven reasoning layer with classically certified holonic execution \citep{lim2024llmmas, ashfaq2025holonicllm} are an active area of work, but the question of where to draw the boundary --- what reasoning to delegate to the agent, what to keep classical --- is itself unsolved.

\section{The Factory as a Cognitive Ecosystem: A Worked Example}
\label{sec:ecosystem}

To make the paradigm concrete, consider a near-future contract manufacturer producing custom medical device housings via injection molding and post-process machining. We describe how an Agent Manufacturing instantiation might handle a single incoming order, anchoring the example in real current deployments where they exist and explicitly noting where the example is a near-future composite.

A customer submits a request via a web interface: a housing meeting specified dimensions, biocompatibility class, and surface finish, with delivery in eleven weeks. A \emph{design agent} --- an LLM with CAD and DFM (design-for-manufacturability) tools, of the kind currently being deployed in Siemens Industrial Copilots integrated with Teamcenter and NX, and in PepsiCo's NVIDIA Omniverse--based factory design pipeline --- interprets the request, retrieves a similar prior part from organizational memory, and proposes a parametric design. It flags two manufacturability concerns (a draft angle below the recommended threshold; a wall-thickness transition likely to cause sink) and offers two alternative geometries. A human design engineer reviews and approves one. This is \emph{interpretive cognition}, and it is the most mature category in current deployments. Even so, reliable DFM reasoning across material and process combinations remains an open problem.

A \emph{process agent} selects materials, mold geometry, and machining strategy, generating a process plan that includes cycle time estimates and tolerance budgets. It queries a \emph{scheduling agent}, which holds a model of current shop loading, to determine when the part can realistically be produced. The scheduling agent negotiates with several \emph{machine agents} representing specific presses and CNC cells; the negotiation surfaces a conflict (the most suitable press is committed to another job during the available window) and resolves it through a renegotiation in which one job is split across two machines. This is \emph{allocative} and \emph{negotiative cognition}. The allocative layer is where classical contract-net MAS solutions already exist \citep{shen2006mas}; the addition Agent Manufacturing makes is that the agents can also reason about why a conflict arose and propose structurally different alternatives, not only bid within a fixed protocol. The negotiative layer beyond simple bidding remains research; the example is a composite.

During production, a \emph{quality agent} monitors in-process measurements. On the third short-run, it detects a drift in a critical dimension. Rather than triggering a fixed alarm, it reasons over the drift pattern, consults memory of prior similar drifts on this material, hypothesizes a likely cause (mold temperature instability following a chiller service the prior week), and routes a request to a \emph{maintenance agent} to verify. This is \emph{diagnostic cognition}. The vision-language-action models that would underpin much of the perception side here --- RT-2 \citep{brohan2023}, OpenVLA \citep{kim2024openvla}, $\pi_0$ \citep{black2024pi0} --- currently achieve, by way of concrete reference, approximately 60\% success on simple grasping tasks after industrial fine-tuning and exhibit positional errors of up to 2.2~cm and 12.4$^\circ$ on high-precision placing \citep{li2025vlatransfer}. The diagnostic reasoning over such inputs is correspondingly limited.

Throughout, human supervisors retain \emph{governance cognition}: setting the goals against which agents are evaluated, approving any action whose risk exceeds a threshold, and auditing decisions after the fact. The factory is not autonomous in the sense of operating without humans; it is autonomous in the sense that coordinative cognition no longer flows primarily through human channels.

Two features of this example deserve emphasis. First, no individual capability is fully present in deployed industrial systems today; the example is a near-future composite, and the most recent systematic survey of foundation-model-based agents in industrial automation places 75\% of reported systems at TRL 4--6 with deployment-oriented evidence remaining at 9.1\% \citep{henkel2026fmagents}. Second, the \emph{coordination structure} --- agents negotiating, reasoning over open-ended state, invoking heterogeneous tools --- is qualitatively different from how factories currently coordinate, even highly automated ones. This is the structural change that makes Agent Manufacturing a distinct paradigm rather than an extension.

The framing of the factory as a ``cognitive ecosystem'' is intended as a structural claim, not a metaphysical one. Specifically, the claim is that the unit of cognitive analysis is no longer the individual engineer or the individual control loop but the network of humans, agents, and artifacts that jointly perform the coordinative work of production. This framing has antecedents in the distributed cognition literature, particularly Hutchins's analysis of ship navigation \citep{hutchins1995} and its identification of three kinds of distribution: across members of a working group, between internal and external representational structures, and across time through accumulated procedures and artifacts. Agent Manufacturing extends Hutchins's framework in one specific way: some of the agents in the distributed cognitive system are now themselves capable of open-vocabulary reasoning, rather than being passive artifacts or narrow tools.

A factory operating in this mode acquires capabilities that workflow-centric automation cannot easily match: faster reconfiguration under disturbance (Siemens's PepsiCo deployment reports a 20\% throughput increase and 10--15\% capital-expenditure reduction from agent-driven simulation-validated reconfiguration), distributed optimization that does not require centralized rescheduling, and the ability to absorb the kinds of high-variety low-volume work that strain conventional MES architectures. It also acquires failure modes that workflow-centric automation does not have: emergent coordination instability when agents pursue locally rational choices that compose badly, opacity in decision chains that span multiple agents and multiple time scales, and the possibility that cognitive concentration shifts from in-house engineering to the providers of the underlying foundation models. The paradigm's appeal and its risks are two sides of the same architectural choice, and we return to both in the sections that follow.

\section{The Labor Question: Coordinative Cognition Is Different}
\label{sec:labor}

Every industrial revolution restructures labor, and discussions of AI and work are now numerous. We make a narrower and more specific claim: Agent Manufacturing displaces a layer of work that has structurally absorbed displaced workers in every prior industrial transition, and there is no obvious next layer to absorb them.

Mechanization displaced artisanal craft but expanded operative and supervisory work. Automation displaced operatives but expanded engineering, planning, and coordinative work --- the layer of process engineers, schedulers, line leaders, supply chain coordinators, and quality engineers that constitutes a large fraction of middle-skill industrial employment in advanced economies. Smart Manufacturing expanded this layer further by adding data engineering and analytics roles. The pattern across two centuries has been remarkably consistent: each wave of automation eliminated tasks at one level of the cognitive hierarchy and created tasks at the level above it.

Coordinative cognition is precisely what Agent Manufacturing targets, and there is no obvious level above it within the manufacturing context. This does not mean coordinative roles disappear; it means their content shifts from performing coordination to \emph{governing} coordination performed by agents --- setting objectives, calibrating trust, intervening on edge cases, auditing decisions.

Two recent findings from labor economics sharpen the concern. First, in their task-based framework, \citet{acemoglu2018ai, acemoglu2022tasks} show that the labor-market effect of automation depends on the balance between a displacement effect (machines replacing labor in particular tasks) and a productivity / reinstatement effect (new tasks being created in which labor has a comparative advantage). When the displacement effect dominates, automation reduces the share of national income going to labor. Second, \citet{acemoglu2024rent} show that automation does not displace tasks at random: it targets \emph{high-rent} tasks --- tasks for which workers were paid wages above their outside option, often because the tasks involved firm-specific knowledge, coordination skill, or interpretive judgment. They estimate that automation accounts for 52\% of the increase in between-group US wage inequality since 1980, and that rent dissipation reduces or even negates the productivity gains from automation in the affected groups. Coordinative cognition in manufacturing is paradigmatically high-rent labor: process engineers, supply-chain coordinators, and production planners earn premia precisely because the work requires accumulated firm-specific judgment. The Acemoglu--Restrepo framework predicts that automating it will be both displacing and rent-dissipating.

\citet{gupta2026agentic} extend the Acemoglu--Restrepo framework specifically to agentic AI, arguing that systems capable of completing multi-step workflows rather than discrete subtasks generate widespread but moderate displacement risk across white-collar roles, rather than immediate mass displacement. Whether this characterization holds across industrial settings remains to be tested, but it identifies the right level at which to ask the question for Agent Manufacturing --- and its prediction of \emph{widespread moderate} rather than \emph{concentrated severe} displacement matches what one would expect when the affected layer is coordinative cognition distributed across many middle-skill industrial roles.

The historical resistance to mechanization --- most famously the Luddites --- is sometimes read as opposition to technology per se. Recent historiography reads it more carefully as resistance to a specific distribution of the gains from mechanization, in a context where the workers who bore the costs had no political mechanism for capturing any of the gains \citep{acemoglu2023power}. The same framing applies here. The substantive question is not whether industrial agents will be deployed (they will, where the economics work) but how authority, accountability, and the productivity gains are distributed across the firms, workers, and societies that bear the transition costs.

We flag, without resolving, three specific questions that deserve more careful treatment than they currently receive in the industrial AI literature: (i) what mix of human-agent task allocation produces the best outcomes for product quality and worker development, as distinct from short-run cost; (ii) what reskilling pathway converts a process engineer into an effective agent governance specialist, and at what scale; (iii) whether the apparent productivity gains of agent-augmented manufacturing accrue to firms, to workers, or to platform owners --- a question whose answer depends on market structure, not technology, and to which we now turn.

\section{The Political Economy of Cognitive Manufacturing Infrastructure}
\label{sec:politicaleconomy}

If \S\ref{sec:labor}'s argument is correct, then the locus of strategic industrial power shifts. Historically that power has rested on land, machinery, energy, labor, and --- most recently --- data. Agent Manufacturing introduces a further locus: the foundation models, memory systems, orchestration platforms, and tool-use infrastructures on which industrial agents depend. We will call this layer \emph{cognitive manufacturing infrastructure}.

This layer has properties that the prior loci did not. It exhibits steep returns to scale (training and operating frontier models is currently the province of a small number of organizations). It is software-defined and therefore mobile across jurisdictions in ways that physical manufacturing capacity is not. And it is dual-use in a stronger sense than industrial machinery: the same foundation model that coordinates a factory can coordinate logistics, energy infrastructure, or military production.

There is a deeper resonance with Hayek's \citep{hayek1945} argument about distributed knowledge that deserves attention here, particularly because it cuts against the surface plausibility of cognitive concentration. Hayek argued that the central economic problem is the coordination of dispersed knowledge --- knowledge of particular times, places, and circumstances that cannot be centralized without loss. The market's price system, in his analysis, functions as a decentralized coordination mechanism precisely because no single planner can hold the relevant knowledge. Agent Manufacturing presents an ambiguous case for this argument. On one hand, foundation-model agents could in principle make Hayekian local knowledge more useable, by giving each ``man on the spot'' an interlocutor capable of contextual reasoning. On the other hand, if those agents all depend on a small number of frontier models trained and operated by a small number of providers, then a substantial portion of the world's industrial coordination becomes mediated through cognitive infrastructure controlled by a few firms --- recreating, in software, exactly the centralization problem Hayek identified. Which of these tendencies dominates is not a technical question.

Three concrete trends in 2024--2026 illustrate that this is not a speculative concern. The EU AI Act (Regulation 2024/1689, in force August 2024, with high-risk obligations enforceable from August 2026) explicitly classifies as high-risk those AI systems operating robots, drones, and medical devices, and those constituting safety components of regulated products. Providers of such systems must complete conformity assessments, register systems in the EU database, implement quality management, and activate post-market monitoring; deployers must implement human oversight and retain logs. Penalties reach EUR~15 million or 3\% of global annual turnover. Industrial agent systems will fall squarely within this regime, with the technical standards (e.g., prEN 18286) still being developed. U.S. export controls on advanced chips and on certain model weights have been progressively tightened across multiple revisions since October 2022 --- including the addition of 140 Chinese entities to the Entity List in December 2024, the proposed AI Diffusion Rule of January 2025 (subsequently rescinded), and the August 2025 arrangement permitting Nvidia H20 and AMD MI308 sales to China subject to a 15\% revenue payment to the U.S. government --- making this a domain of unusually volatile, openly geopolitical policy. China's State Council ``AI+'' Action of August 2024 and the joint MIIT ``AI + Manufacturing'' plan target the deployment of 3--5 general-purpose industrial models, 1{,}000 industrial intelligent agents, 100 high-quality manufacturing datasets, and 500 demonstration scenarios by 2027, with explicit emphasis on building a domestic alternative to foreign frontier models.

The geopolitical question this raises is not the familiar one of who manufactures what, but the less familiar one of who controls the cognitive layer through which manufacturing is coordinated. A country with substantial physical manufacturing capacity but dependent on foreign cognitive infrastructure is in a different strategic position than one with both. This is a different framing from ``AI sovereignty'' in the consumer-AI sense, and we suspect it will become more salient as industrial deployments mature.

We do not have policy recommendations to offer. We note only that the existing manufacturing-policy literature is largely written under assumptions --- that industrial capacity is the relevant unit, that data sovereignty is the relevant data concern --- that Agent Manufacturing renders incomplete. Several questions naturally follow that we leave for separate treatment: whether industrial foundation models are inherently monopolistic, how industrial agent memory transfers when a provider relationship ends, and how cognitive-infrastructure concentration differs from the prior cloud-and-ERP concentration that manufacturers already navigate.

\section{Research Agenda}
\label{sec:agenda}

Many of the open problems in foundation-model agents are well-known and discussed in the general agent literature \citep{xi2023agentsurvey, wang2024agentsurvey, chen2024mgassurvey}. We list here a smaller set of problems whose form is specific to manufacturing, in the sense that solving them for, say, web agents would not solve them for industrial agents.

\paragraph{Physical grounding under industrial tolerances.}
Vision-language-action models such as RT-2 \citep{brohan2023}, the Open X-Embodiment collection \citep{oxe2024}, OpenVLA \citep{kim2024openvla}, and $\pi_0$ \citep{black2024pi0} have demonstrated that foundation-model-based control of physical systems is feasible in laboratory settings. Industrial deployment requires reliability and precision that exceed these demonstrations by orders of magnitude. The most recent benchmark we are aware of \citep{li2025vlatransfer} finds that $\pi_0$ after fine-tuning achieves approximately 60\% success rate on simple grasping tasks, with positional error up to 2.2~cm and 12.4$^\circ$ on high-precision placing --- orders of magnitude removed from the micron-level tolerances of injection molding, the six-sigma defect rates of automotive production, or the deterministic timing of process control. The relevant research direction is not larger models alone but tighter coupling between learned controllers and verifiable physical models --- what some recent process-industry work \citep{ren2025industrialfm} calls mechanism-informed industrial foundation models.

\paragraph{Real-time guarantees in agent-based coordination.}
Industrial control systems are designed around deterministic cycle times and bounded latency. Foundation-model agents are stochastic and computationally expensive, and their latency is poorly bounded. Architectures that combine slow, deliberative agent reasoning with fast, deterministic control loops are an active research area but not a solved problem; the question of which decisions can safely be delegated to slow reasoning is itself open. The hybrid holonic-LLM architectures emerging in 2024--2025 \citep{lim2024llmmas, ashfaq2025holonicllm} are one promising direction but lack the safety analysis that would be required for certified deployment.

\paragraph{Safety certification for thick-autonomy systems.}
Functional safety standards (IEC 61508, ISO 13849) assume control logic that can be exhaustively analyzed. Open-vocabulary agent behavior cannot be. Either the standards need to evolve to accommodate statistical safety arguments --- as has begun to happen in autonomous vehicles --- or industrial agent architectures need to expose deterministic sub-components with classically certifiable behavior, with agent reasoning constrained to operate above them. Both routes are open research problems. The EU AI Act's emerging harmonized standards (prEN 18286 entered enquiry in October 2025) will shape practice in this area, though they are not yet a solution.

\paragraph{Organizational memory.}
A factory's cognitive value increasingly resides in the accumulated memory of its agents: lessons from past disturbances, customer-specific quirks, supplier reliability patterns, tribal knowledge made explicit. The question of how such memory should be represented, how it transfers when a system is replaced or extended, who owns it, and how it can be audited has no current good answer. This is a manufacturing-specific instantiation of broader questions about agent memory \citep{park2023generative, sumers2023}, but with the added constraint that industrial memory frequently includes information of competitive or regulatory consequence.

\paragraph{Human-agent governance interfaces.}
The mechanisms by which humans set objectives, calibrate trust, intervene on edge cases, and audit decisions in agent-coordinated factories are largely unspecified. Existing HMI standards address human interaction with deterministic systems; they do not address how a line manager should oversee an agent that is reasoning over open-ended state. This is a problem at the intersection of human factors, AI evaluation, and industrial software engineering, and we are not aware of a coherent body of work yet addressing it.

\paragraph{Engineering education.}
The engineers who will design, deploy, and govern Agent Manufacturing systems are largely in school today, being trained in curricula that assume the prior paradigm. Curricular research on integrating LLM agents, vision-language-action models, and agent governance into manufacturing engineering programs --- without abandoning the physical fundamentals that remain necessary --- is underway in several institutions but is not yet a coherent body of work.

\section{Conclusion}
\label{sec:conclusion}

We have argued that the convergence of foundation models, vision-language-action models, and multi-agent orchestration constitutes the technical substrate for a distinct manufacturing paradigm, which we call Agent Manufacturing. The paradigm is distinct from Smart Manufacturing in that the coordination mechanism, not only the execution layer, is reorganized; distinct from Cognitive Manufacturing in the strength of cognition it claims; distinct from Industry 5.0 in being descriptive rather than normative; and distinct from classical multi-agent manufacturing in the thickness of the autonomy it requires its agents to exhibit.

We have been deliberate about the limits of this argument. The empirical evidence supports the claim that the technology is \emph{of the right type} to displace coordinative cognition, not that current systems have done so. The 9.1\% deployment rate found in the most recent systematic survey, the $\sim$60\% grasping success of current VLAs after industrial fine-tuning, and the 2.2~cm positional error on precision placing tasks are not, on their face, the numbers of an imminent transition. They are, instead, the numbers of a paradigm at TRL 4--6 --- past laboratory curiosity, short of industrial deployment. Whether Agent Manufacturing matures into a deployed paradigm depends on developments that current evidence does not foreclose in either direction.

We should also be explicit about a deeper uncertainty. Foundation models are trained predominantly on internet-scale text and image data; their grounding in physical industrial reality is incidental, not principled. It is possible that the trajectory from current capabilities to industrial-grade coordination will not be reached by scaling current architectures but will require fundamentally different approaches --- mechanism-informed models, neuro-symbolic architectures of the kind being developed in the LLM-ACTR line of work, tighter integration with verified physical simulators, or paradigms not yet in view. Our argument does not depend on which of these paths succeeds. It depends only on the claim that \emph{some} approach within this family will eventually be deployable in the coordinative layer, and that the research, governance, and educational questions raised here will apply when it is.

The technology is not yet ready for unsupervised industrial deployment, and our argument does not require it to be. The argument requires only that the trajectory is now clear enough, and the technical substrate present enough, that the relevant research, governance, and educational questions need to be posed now, in terms specific to manufacturing rather than borrowed from adjacent fields. The cost of posing them late is that they will be answered by default --- by whichever firms happen to deploy first, in whichever jurisdictions happen to regulate first, under whichever distribution of authority happens to emerge from market dynamics alone.

The deeper observation behind this paper is that the boundary between machine execution and human coordination, which has held across four industrial revolutions, is the boundary now being crossed. Whether the result is a manufacturing civilization that retains meaningful human agency or one in which industrial cognition concentrates in a small number of model providers is not a question the technology answers on its own. It is a question of who builds, who governs, and who is included in the transition.

\section*{Use of AI in Preparation of This Manuscript}

The author used generative AI tools --- including ChatGPT (OpenAI) and Claude (Anthropic) --- in the preparation of this manuscript, covering tasks including drafting, structural revision, literature search, and iterative editing across multiple versions. The intellectual direction, central thesis, and final responsibility for all claims, citations, and wording rest with the author. Consistent with arXiv policy and the principle that authorship requires accountability that AI systems cannot bear, no AI tool is listed as an author. Readers and reviewers are encouraged to evaluate this manuscript on the merit of its arguments and the accuracy of its citations; the author welcomes correction of any errors that may have been introduced during AI-assisted drafting.

\bibliography{references}

\end{document}